# UPGRADE OF A ROBOT WORKSTATION FOR POSITIONING OF MEASURING OBJECTS ON CMM

**Dipl.-Ing. Deni Šabić**
**Mr.sci. Damir Štrbac**
**Doc. dr. sci Samir Lemeš**
**Doc. dr. sci. Malik Čabaravdić**
**University of Zenica, Faculty of Mechanical Engineering,**
**Fakultetska 1, Zenica**
**Bosnia&Herzegovina**

**ABSTRACT**
*In order to decrease the measuring cycle time on the coordinate measuring machine (CMM) a robot workstation for the positioning of measuring objects was created. The application of a simple 5-axis industrial robot enables the positioning of the objects within the working space of CMM and measuring of different surfaces on the same object without human intervention. In this article an upgrade of an existing robot workstation through different design measures is shown. The main goal of this upgrade is to improve the measuring accuracy of the complex robot-CMM system.*

## 1. INTRODUCTION

Coordinate measuring machines as an element of dimensional metrology are very important link in the chain of precise measurement and control. Coordinate metrology and coordinate measuring machines are very young operating systems. They are in use since the early sixties of the 20th century, but very little is known about them compared to other technologies. Intensive development of information technology in recent years has an increasing effect on the development of coordinate metrology and evaluation of coordinate measuring machines. These measuring machines generally define tolerances of dimensional characteristics, shape and position of the actual workpieces in relation to designed ones. A real picture of the workpiece and its deviation from the designed model of a given piece are obtained by measuring. CMM is a measuring machine with high speed measurements in which the positioning and rotation of the measured object is done manually within the working area of the measuring machine.

In order to decrease the measuring cycle time on the coordinate measuring machine (CMM) a robot workstation for the positioning of measuring objects was created [1]. The application of a simple 5-axis industrial robot enables the positioning of the objects within the working space of CMM and measuring of different surfaces on the same object without human intervention.

The idea of coupling of CMM with an industrial robot was not treated by many authors. Most similar research work was done by J. Santolaria and J.J Aguilar, Universidad de Zaragoza [2]. In this work the priority was given to the development of kinematic chain model of robot manipulators and CMM. Many research works are mainly focused to the impact of diverse factors on the measurement uncertainty of the CMM machines [3,4,5,6].

The first experiments in the above mentioned CMM-robot workstation showed that it was possible to conduct the measurements in this complex system, but the measurement precision was lower than expected [1]. Therefore, the existing work cell was upgraded by an improved construction design, and the results of this improvement are shown in this paper.

## 2. NEW WORKCELL SETUP

In first experiments the base of the robot was mounted to the base of the CMM. The movement of the CMM's frame during the measurement caused vibrations and these vibrations were transferred to the robot and whole measuring system.

The main idea in a new setup of the work cell was to detach the base of the robot from CMM. With help of a new construction (see Figure 1) the robot base was attached and anchored directly to the floor, decreasing the influence of the CMM's vibrations on the measurements.

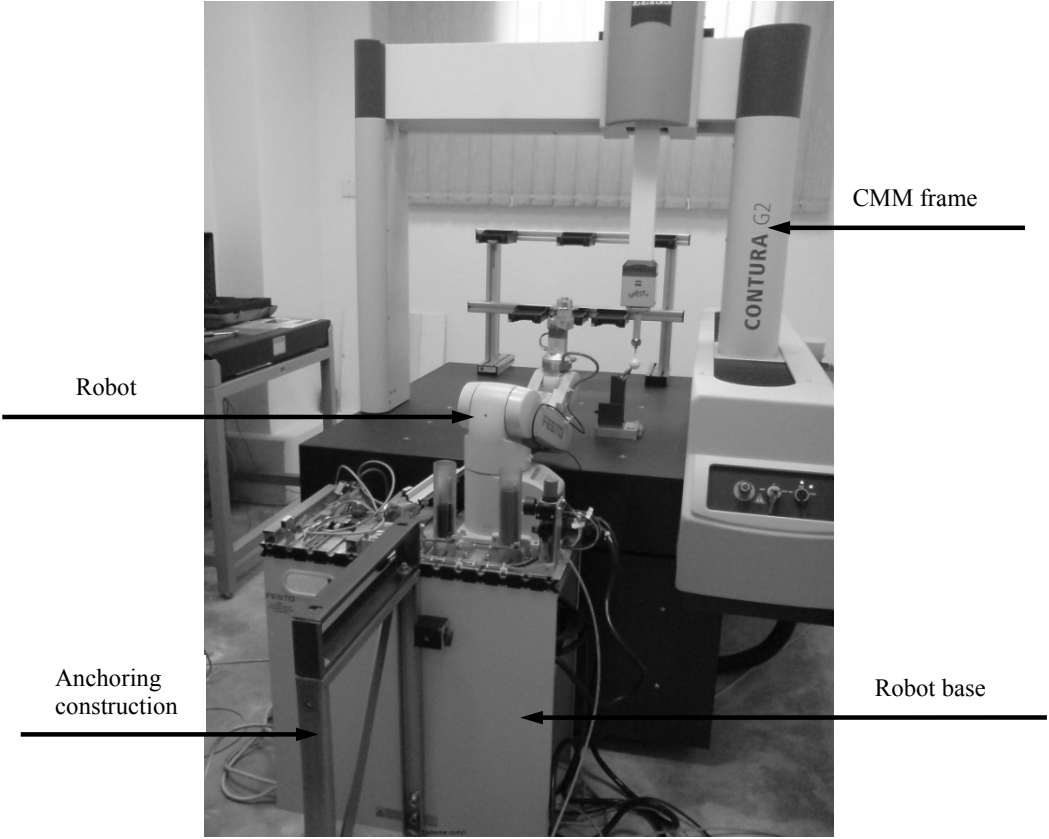

*Figure 1. New anchoring construction for the robot*

## 3. DESIGN OF EXPERIMENT

For practical measurement was used a complex measuring system consisting of Contura G2 CMM Active coordinate measuring machine with additional measuring instruments (tentacles, calibrating tools, etc.) and 5-axis educational robot Mitsubishi RV-2AJ with equipment for gripping the measuring objects. The equipment is situated in a closed laboratory space with nearly constant temperature and humidity. Due to the closure of the laboratory space and because of the very small flow of people and materials in the workplace it is assumed that the impact of dust to the measurement is not significant. In order to consider different geometrical shapes during the measurement the experiment was focused on four geometric planes: a-frontal plane of the cylinder, b-cone on the cylinder, c-side conical plane of the cylinder and d-frontal plane of the larger cylinder (Figure 2).

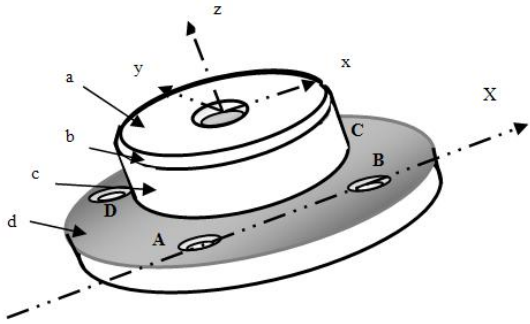

*Figure 2. Measuring object with defined geometrical measuring planes [1]*

If the centers of the circular holes on the measuring object are assigned as the points A, B, C and D (Figure 2), then the line passing through the points A and B defines the X axis (in the plane d), which is used for orientation of the tentacle. The measured dimensions for the experiment are defined on the measuring object: diameter d1, diameter d2, diameter d3 and height H (Figure3a)). These geometrical are related to: xy plane, plane ABCD, conical plane 1 and conical plane 2 (Figure 3b)).

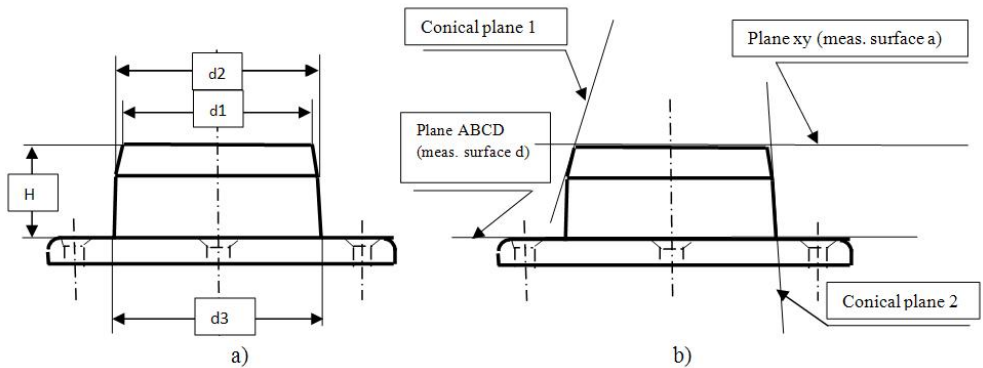

*Figure 3. Measured dimensions a) and planes related to those dimensions[1]*

In order to ensure the accuracy of the measurements and get the correct measurement results, the calibration of the measuring tool on CMM and the examination of the whole measurement system was performed before the measurement process.For all defined geometrical planes and mesasured dimensions 10 dimensional measurements was completed after the calibration. On the basis of these results the measurement uncertainty as a measure of the validity of the measuring results was determined.

Position the measuring object is defined by the position of the robot arm in the CMM's coordinate space. Between each measurement robot arm was moved from measuring position to another position in the space and then moved back. All positions of the robot are stored during the off-line teaching phase.

## 4. RESULT ANALISYS

After performing of 10 measurements in the improved measuring cell average value, standard deviation and coefficient of variation for each measured dimension were calculated. These results were compared with the results from the previous experiments with old work cell setup. The results are given in table 1.

*Table 1. Comparison of the measurements from previous and improved work cell*

|  | The measured values of the geometrical size | | | | | | | |
|---|---|---|---|---|---|---|---|---|
|  | Diameter d1 (previous) | **Diameter d1 (improved)** | Diameter d2 (previous) | **Diameter d2 (improved)** | Diameter d3 (previous) | **Diameter d3 (improved)** | Height H (previous) | **Height H (improved)** |
| Nominal value | 39,00000 | **39,00000** | 42,50000 | **42,50000** | 42,90000 | **42,90000** | 18,70000 | **18,70000** |
| Average value | 39,03030 | **39,00817** | 42,51910 | **42,50757** | 42,91840 | **42,89127** | 18,82070 | **18,77220** |
| Standard deviation $\sigma 1$ | 0,00600 | **0,00049** | 0,00450 | **0,00019** | 0,00470 | **0,00025** | 0,00360 | **0,00011** |
| Coefficient of variation V1 | 0,01540 | **0,00126** | 0,01070 | **0,00045** | 0,01090 | **0,00058** | 0,01930 | **0,00058** |

As from the Table 1 can be seen the results in the improved work cell are much better than the results in the previous cell. Standard variation and coefficient of variation were up to 10 times decreased and the quality of measurement with the new work cell setup was significantly improved.

## 5. CONCLUSION

In order to increase the degree of automation by the measuring process with coordinate measuring machines, an industrial robot for positioning of measuring objects was used. The measurement of the objects with complicated geometries requires often manual repositioning, including redefinition of the local coordinate system. If an industrial robot issued to manipulate the measured object, such a process could be accelerated.

The measurements of the dimensions of the measured object were conducted by complex CMM-robot measuring system, with movements performed between each single measurement. With the help of a new construction for anchoring of the robot base the precision of the measurement was significantly improved because of better dumping of vibrations that were obvious during every movement phase between measurements.

## 6. ACKNOWLEDGEMENTS

This research was supported in part by the Ministry for Education and Science of the Federation of Bosnia and Herzegovina.